\documentclass[sigconf]{acmart}
\usepackage{multirow}
\usepackage{pifont}
\usepackage{graphicx}
\usepackage{amsmath}
\usepackage{amsthm}
\usepackage{booktabs}
\usepackage{algorithm}
\usepackage{algorithmic}
\AtBeginDocument{%
  }

\setcopyright{acmlicensed}

\copyrightyear{2023}
\acmYear{2023}

\acmConference[MM '23]{Proceedings of the 31st ACM International Conference on Multimedia}{October 29--November 3, 2023}{Ottawa, ON, Canada}

\acmBooktitle{Proceedings of the 31st ACM International Conference on Multimedia (MM '23), October 29--November 3, 2023, Ottawa, ON, Canada}
\acmPrice{15.00}
\acmISBN{979-8-4007-0108-5/23/10}
\acmDOI{10.1145/3581783.3612090}
\begin{document}

\title{Object Detection Difficulty: Suppressing Over-aggregation for Faster and Better Video Object Detection}

\author{Bingqing Zhang}
\affiliation{%
  \institution{Renmin University of China}
  \streetaddress{No. 59, Zhongguancun Street, Haidian Dist.}
  \city{Beijing}
  \country{China}
  \postcode{100872}
}
\email{bingqing.zhang@ruc.edu.cn}

\author{Sen Wang}
\affiliation{%
  \institution{The University of Queensland}
  \streetaddress{The University of Queensland, Sir Fred Schonell Dr, St Lucia QLD 4072}
  \city{Brisbane}
  \country{Australia}
  }
\email{sen.wang@uq.edu.au}

\author{Yifan Liu}
\affiliation{%
  \institution{The University of Adelaide}
  \city{Adelaide}
  \country{Australia}
}
\email{yifanliu0926@gmail.com}

\author{Brano Kusy}
\affiliation{%
 \institution{Data 61, CSIRO}
 \city{Brisbane}
 \country{Australia}}
\email{brano.kusy@data61.csiro.au}

\author{Xue Li}
\affiliation{%
  \institution{The University of Queensland}
  \city{Brisbane}
  \country{Australia}}
\email{xueli@itee.uq.edu.au}

\author{Jiajun Liu}
\authornote{Corresponding author}
\affiliation{%
  \institution{Data 61, CSIRO}
  \city{Brisbane}
  \country{Australia}}
\email{jiajun.liu@csiro.au}

\renewcommand{\shortauthors}{Bingqing Zhang et al.}

\newcommand\minisection[1]{\vspace{2mm}\noindent \textbf{#1}}

\begin{abstract}
Current video object detection (VOD) models often encounter issues with over-aggregation due to redundant aggregation strategies, which perform feature aggregation on every frame. This results in suboptimal performance and increased computational complexity. In this work, we propose an image-level Object Detection Difficulty (ODD) metric to quantify the difficulty of detecting objects in a given image. The derived ODD scores can be used in the VOD process to mitigate over-aggregation. Specifically, we train an ODD predictor as an auxiliary head of a still-image object detector to compute the ODD score for each image based on the discrepancies between detection results and ground-truth bounding boxes. The ODD score enhances the VOD system in two ways: 1) it enables the VOD system to select superior global reference frames, thereby improving overall accuracy; and 2) it serves as an indicator in the newly designed ODD Scheduler to eliminate the aggregation of frames that are easy to detect, thus accelerating the VOD process. Comprehensive experiments demonstrate that, when utilized for selecting global reference frames, ODD-VOD consistently enhances the accuracy of Global-frame-based VOD models. When employed for acceleration, ODD-VOD consistently improves the frames per second (FPS) by an average of $73.3\%$ across 8 different VOD models without sacrificing accuracy. When combined, ODD-VOD attains state-of-the-art performance when competing with many VOD methods in both accuracy and speed. Our work represents a significant advancement towards making VOD more practical for real-world applications. The code will be released at https://github.com/bingqingzhang/odd-vod.
\end{abstract}




\maketitle

\section{Introduction}

Video Object Detection (VOD) \cite{fgfa, selsa, dsfnet} focuses on identifying objects within videos by utilizing rich spatial and temporal data. This task holds significant importance in the field of multimedia. VOD models typically build upon the success of modern Still Image Object Detectors (SIODs) \cite{faster_rcnn, yolov3, detr, deformable_detr}, sampling a series of reference frames and aggregating them to support the frame being processed. This approach has proven effective for enhancing frame feature representation and improving object detection performance. Consequently, many state-of-the-art models focus on designing aggregation modules, such as SELSA \cite{selsa} and LRM in MEGA \cite{mega}. However, feature aggregation operations applied to every frame can introduce high computational costs and decrease detection speed, which ultimately limits the practical applicability of VODs in real-life applications. We refer to this as the \textit{over-aggregation} problem. Moreover, low-quality reference frames may not provide any benefits in the VOD aggregation step due to their limited information content, resulting in suboptimal performance.

\begin{figure}[tbp]
    \centering
    \includegraphics[width=0.5\textwidth]{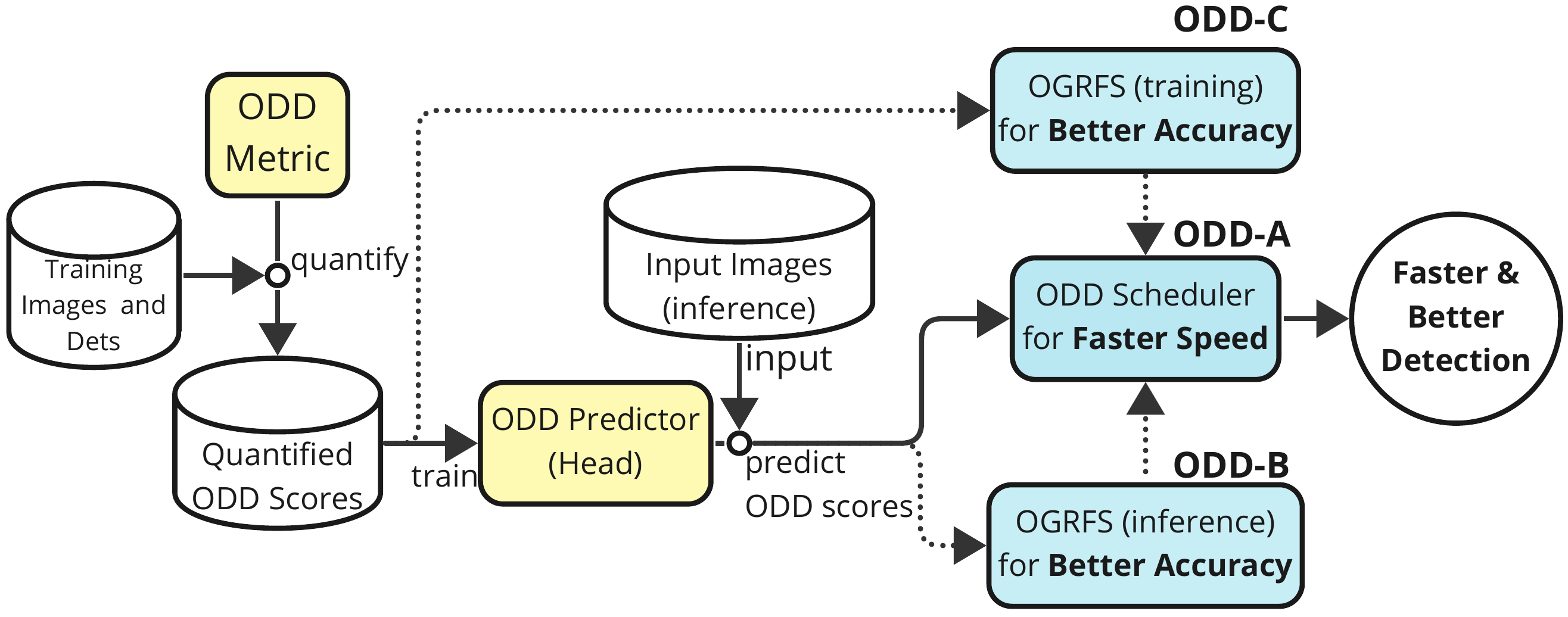}
    \caption{An overview of ODD-VOD. The dotted lines represent optional operations. There are five ODD-based components: ODD Metric, ODD Predictor for predicting ODD scores of testing frames. ODD Scheduler (ODD-A) switches between VOD and SIOD to accelerate the inference speed. ODD-based Global Reference Frame Selector (OGRFS) selects low-ODD-score frames (easy to detect) as global reference frames in the training stage (ODD-C) and inference stage (ODD-B) for better accuracy. The ODD-* labels are created for easier reference in the rest of the paper.}
    \label{fig:overview}
\end{figure}

Methods for addressing the speed-accuracy trade-off in VODs have been proposed and can be categorized into two types: plug-in and unified methods. Plug-in methods \cite{dfa} are employed alongside existing VOD models, while unified methods \cite{Zhu2017TowardsHP, d_or_t, query_prop} combine two distinct strategies, aggregation and propagation, within the VOD algorithm. In key frames, unified methods use feature aggregation to improve detection accuracy, while in non-key frames, a propagation module updates motion information. Although unified methods alleviate the over-aggregation problem, their key-frame selection strategy or scheduler is overly simplistic, relying on either interval-based \cite{Zhu2017TowardsHP} or heuristic sampling \cite{d_or_t}. Meanwhile, the quality of reference frames remains largely unexplored in VOD literature.

In this paper, we present an efficient VOD framework called Object Detection Difficulty-based VOD (ODD-VOD) to address the over-aggregation problem and enhance detection accuracy. The underlying intuition is twofold: first, feature aggregation is only necessary for frames that pose \textit{difficulties} for SIODs; second, feature aggregation yields the most significant improvements over SIODs when the aggregated images are of high quality and easily detected by SIODs. To quantify the difficulty of an image for SIODs, we propose the ODD metric, which measures a ground-truth ODD score given a training image, its annotations, and its detection results from a pre-trained image detector. We then train an ODD Predictor that supports the ODD-VOD pipeline (Figure \ref{fig:overview}). The main technical contributions of our proposed framework are as follows:
\begin{enumerate}
    \item \textbf{ODD Metric}: We formulate a non-trivial metric to quantify the ODD score for an image.
    \item \textbf{ODD Predictor}: We first train an ODD Predictor module using quantified ODD scores to measure the detection difficulty of testing frames.
    \item \textbf{ODD-A}: We then propose an ODD
Scheduler that switches between VOD and SIOD for a given input frame to increase inference speed.
    \item \textbf{ODD-B/C}: We present ODD-based Global Reference Frame Selector (OGRFS). As a plug-in method, OGRFS utilizes ODD scores to guide the selection of global reference frames, thereby improving the quality of aggregated features. 
\end{enumerate}
OGRFS can be utilized in two ways. It can be directly applied during inference without retraining the VOD model, resulting in improved detection accuracy (ODD-B). To further enhance detection accuracy, OGRFS can be employed during both training and inference stages (ODD-B+C).

\begin{figure}[tbp]
    \centering
    \includegraphics[width=0.48\textwidth]{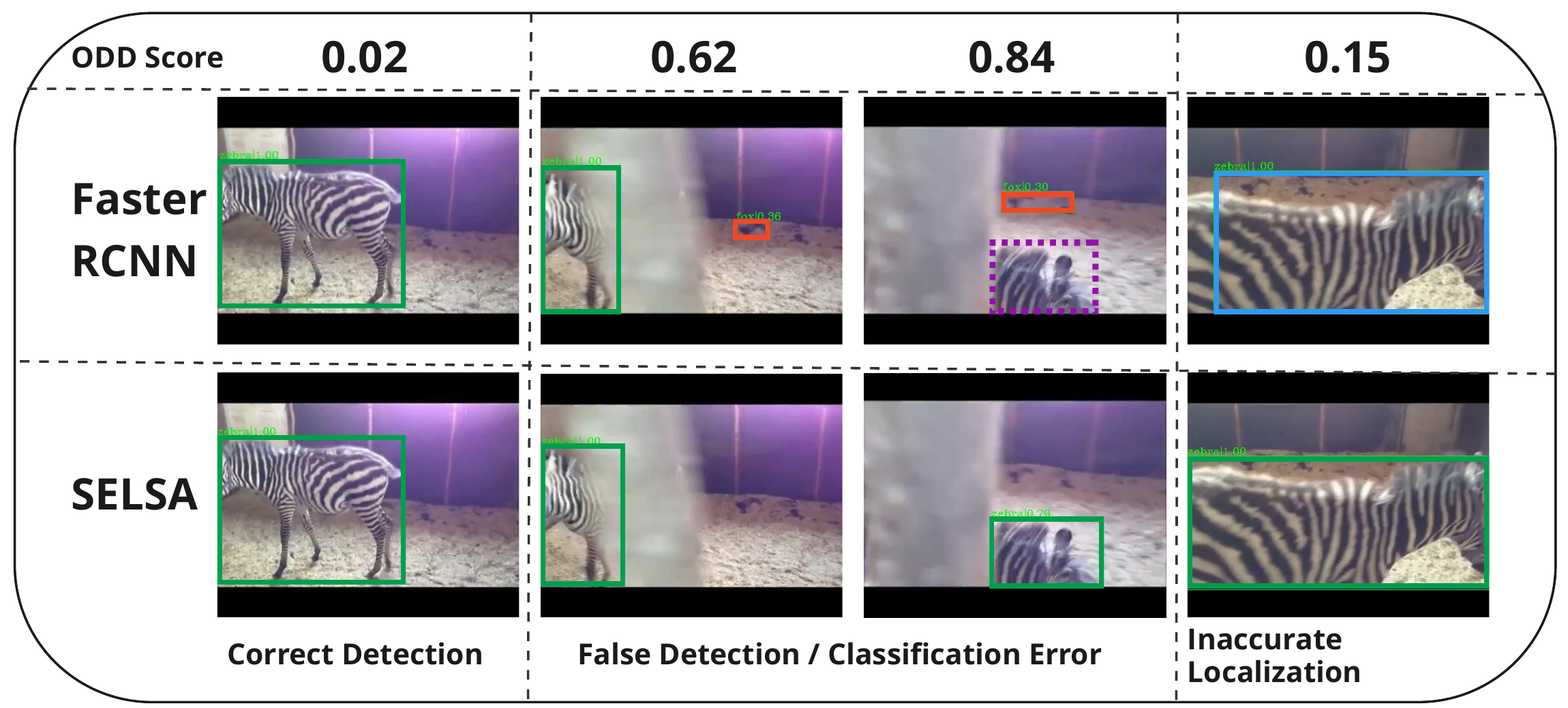}
    \caption{The predicted ODD scores have an inverse correlation with detection quality on two detectors. If the ODD score of a frame is low, the SIOD (Faster R-CNN) will have the same detection results as the video detector (SELSA). As the ODD score increases (making detection more difficult), the VOD will have significantly better results.}
    \label{fig:vis_intro}
\end{figure}

Figure \ref{fig:vis_intro} illustrates how the proposed ODD score captures the difficulty levels for an SIOD and how this subsequently reflects the benefits of VOD/feature aggregation. We display the quantified ODD scores and the outputs of two detectors on a test video snippet. There is a strong inverse correlation between the predicted ODD scores and the accuracy of detection results from Faster R-CNN \cite{faster_rcnn} (a typical SIOD). When the ODD scores are low (easy to detect), there is little difference between Faster R-CNN and SELSA \cite{selsa} (VOD). As the ODD score increases (becoming harder to detect due to occlusion, blurriness and so on), Faster R-CNN starts to produce inaccurate object localization (blue box) and eventually misclassifies the object (red boxes) or even misses detections (purple box) (see more cases in Section \ref{sec:visualization}). In our ODD-VOD, detection accuracy and speed can be simultaneously optimized using ODD scores. When a frame has a low ODD score, an SIOD can quickly process it without sacrificing accuracy and bypass VOD/feature aggregation. When the ODD score is higher, we can switch to VOD to detect these frames, maximizing aggregation benefits; furthermore, in the aggregation process, we use reference frames with low ODD scores for improved results.

In the Experiments section, we present extensive empirical results demonstrating that our proposed method can achieve significant speed improvements and enhanced accuracy across a range of widely-used VOD models.

\section{Related Work}
\minisection{Object Detection in Still Images}
As one of the most critical computer vision tasks, object detection has been extensively studied in academia and industry. Detectors in still images can be classified into two types: two-stage and one-stage. Two-stage detectors first extract objects from background areas and then determine their categories and positions \cite{fast_rcnn,faster_rcnn,r_fcn,cascade_rcnn,dynamic_rcnn}. On the other hand, one-stage detectors \cite{yolov3,ssd,yolox,yolov5,yolov7} can be faster than the two-stage ones. These detectors regard the detection task as a regression problem to obtain the types and positions of bounding boxes. Recent years, applying the transformer \cite{attention_is_all_you_need} to object detection is becoming a new research hotspot \cite{detr,meng2021-CondDETR,liu2022dabdetr, dino}.

\minisection{Object Detection in Videos}
Unlike object detection in still images, video object detectors utilize spatial and temporal information in frames. Based on the different types of reference frames, existing methods can be classified into three categories: local-frame-based, global-frame-based, and both-frame-based.
Local-frame-based methods aggregate temporal information of nearby frames \cite{dff,fgfa,rdn,iffnet,stsn}. Global-frame-based methods select a range of reference frames in the whole videos and seek to enhance the features with semantic information \cite{Shvets2019LeveragingLT,selsa,temporal_roi_align,afa,tfblender,hvr_net,dsfnet}. Apart from the cite reference frames from the local or global position, MEGA \cite{mega} samples frames from both positions to obtain semantic information and motion information. However, aggregating on each frame may cause the over-aggregation problem, which brings high computational costs.

Therefore, efficient VODs are designed to reduce aggregation costs. Plug-in efficient methods \cite{dfa} can be attached to other VODs for decreasing reference frames. Unifed-based efficient methods \cite{Zhu2017TowardsHP,d_or_t,query_prop, psla, eovod} combine dense (aggregation) and sparse (propagation) detectors to achieve faster speed. However, propagation between frames is vulnerable and sensitive to changes in objects' appearance and positions. Our method replaces the sparse detector with an SIOD to avoid the object motion problem.

\begin{figure}[tbp]
    \centering
    \includegraphics[width=0.48\textwidth]{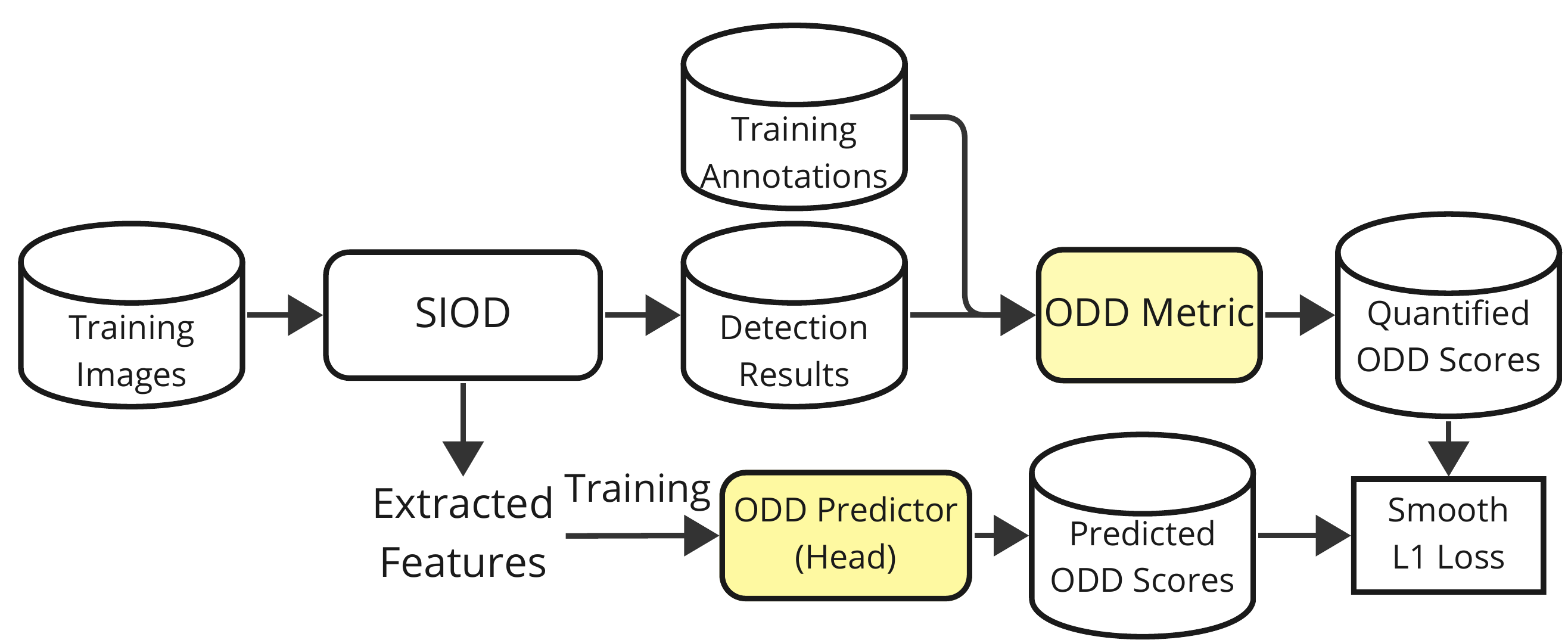}
    \caption{Training Pipeline for ODD Predictor. It takes four steps: 1) training an SIOD; 2) obtaining detection results of training images; 3) quantifying ODD scores; 4) using training images and ODD scores to train the ODD Predictor.}
    \label{fig:odd_predictor}
\end{figure}

\section{Method}
We first formally define the concept of ODD, and describe the process to quantify the ODD scores from the training set. We then present the details of ODD-VOD and its components.

\subsection{Definition of Object Detection Difficulty}
\label{sec:odd_defi}
The most prevalent metric for evaluating the performance of various object detectors is the Mean Average Precision (mAP). However, mAP is a dataset-level metric, which makes it unsuitable for directly assessing the detection results on individual images. To address this limitation, we introduce the concept of Object Detection Difficulty (ODD), an image-level metric for object detection performance.

We propose an image-level metric for object detection difficulty, derived from the results of a SIOD model and the ground-truth bounding boxes in the training set. This image-level metric serves as the ground-truth signal for training an ODD Predictor, which can then estimate an ODD score for each image, making it a valuable tool for training or inference.

After obtaining the detection results from the SIOD model and the ground-truth bounding boxes for an image in the training set, we classify each predicted bounding box into four categories: positive, negative, near-positive, and multi-positive. If a bounding box has the maximum Intersection over Union (IoU) with a ground-truth bounding box, and the IoU exceeds the positive threshold ($t_1$), we label this result as positive. If a bounding box's IoU is greater than $t_1$ but not the maximum, we classify this result as multi-positive. If a bounding box's IoU does not reach the positive threshold but is numerically close (e.g., $t_1$ is 0.5 and the IoU is 0.49), we designate this result as a near-positive sample. The concept of near-positive samples is inspired by the sampling strategy of Region Proposal Network (RPN) \cite{faster_rcnn}, which assigns positive samples in the second stage. Furthermore, the IoU of a near-positive sample should be larger than the near-positive threshold ($t_2$). Finally, if the IoU of a predicted bounding box is smaller than $t_2$, we categorize this result as negative. Among these categories, positive, near-positive, and multi-positive results contribute positively to the ground-truth ODD score.

Then, we can use a unified formula to define the weighted sample ($ws$). It can be written as:
\begin{flalign}
\label{eq:ws}
ws(p) &= \nonumber \sum_{l \in L}\sum_{max(IoU_i)}C_{l, i}\mathbf{1_P}(p)   +  \sum_{l \in L} \sum_{IoU_i \in \left [ t_2, t_1 \right )} \frac{1}{2} C_{l, i} \mathbf{1_{NR}}(p) &&\\  &+ \sum_{l \in L} \sum_{IoU_i \in \left [ t_1, 1 \right ]} C_{l, i} \mathbf{1_M}(p) + \sum_{l \in L} \sum_{IoU_i \in \left [0, t_2 \right )} C_{l, i} \mathbf{1_N}(p),&&
\end{flalign}

where $L$ is the label of different objects, $C$ is the confidence score of the detection result, $t_1$ is the positive threshold and $t_2$ is the near-positive threshold which is mentioned above. $\mathbf{1_P}$, $\mathbf{1_{NR}}$, $\mathbf{1_M}$ and $\mathbf{1_N}$ are indicator functions, which are used to determine whether the current result belongs to the corresponding sample (positive, near-positive, multi-positive and negative sample correspondingly). The output of this function has two values, 0 and 1. In addition, there is a parameter $p$ in equation \ref{eq:ws}. When $p$ is equal to 1, $\mathbf{1_P}$, $\mathbf{1_{NR}}$ and $\mathbf{1_M}$ will work to find all positive samples. And when $p$ is 0, $ws$ will calculate weighted negative samples. The weighted sample is an important measurement to calculate the output of a detector on one image. Borrowing the idea of the F1-score, we also use harmonic means to balance different samples.

\noindent First, the weighted precision ($wp$) can be defined as:
\begin{equation}
    wp = \left\{ \begin{matrix}
        
    \frac{ws(p=1)}{ws(p=1) + ws(p=0) } \\ \\ 1,   \textrm{if no gt bbox}
\end{matrix} \right. 
\end{equation}
 Here the denominator cannot be 0 due to the structure of object detectors. And the weighted recall ($wr$) can be defined as:
\begin{equation}
    wr = \left\{\begin{matrix}
\frac{ws(p=1)}{max(\textrm{total\_gt\_sample}, ws(p=1))} \\ \\
1,   \textrm{if no gt bbox}
\end{matrix}\right. ,
\end{equation}
where $\textrm{total\_gt\_sample}$ is the total number of ground truth proposals in one image. Finally, we can define the ODD with $wp$ and $wr$:
\begin{equation}
    ODD = 1 - 2 \cdot \frac{wp \times wr}{wp + wr + \varepsilon } ,
    \label{eq:odd}
\end{equation}
where $\varepsilon$ is a tiny value to prevent the denominator from being 0. The value range of ODD is between 0 and 1. If the ODD score is high, it means the current image is hard to detect for the object detector and 
vice versa.

\subsection{ODD-VOD Framework}
An overview of the ODD-VOD framework is presented in Figure \ref{fig:overview}. The basic input of the framework is the quantified ODD scores (ODD ground truth scores), which can be calculated from an SIOD and the training dataset. ODD Predictor is used to predict ODD scores of testing frames, which is an important input for the ODD Scheduler. The ODD Scheduler (ODD-A) is deployed in a hybrid detection pipeline for faster speed. OGRFS can be used in the training stage (ODD-C) and inference stage (ODD-B), which select lowest $k$ ODD score global reference images for better detection accuracy. We now introduce the ODD-based components in detail.

\begin{figure}[tbp]
    \centering
    \includegraphics[width=0.48\textwidth]{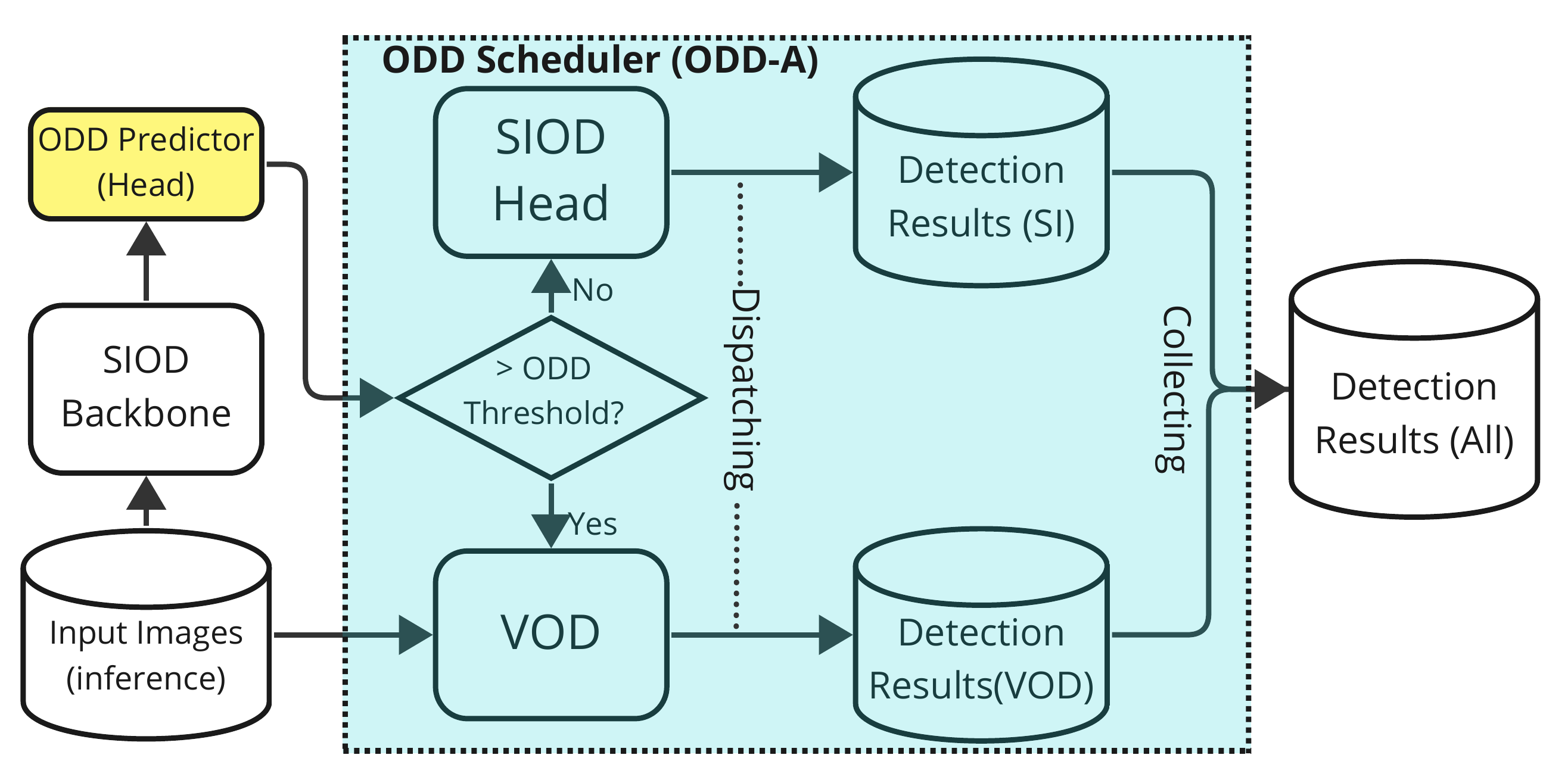}
    \caption{ODD Scheduler for faster detection. The input is predicted ODD scores from ODD Predictor. Then, the ODD Scheduler will use the ODD score to determine which detector to detect the current frame (Dispatching). After generating results, ODD Scheduler will collect them together (Collecting). In addition, ODD-B (OGRFS in inference) can be used to improve detection accuracy.}
    \label{fig:odd_scheduler}
\end{figure}

\minisection{Training the ODD Predictor}


The ODD Predictor is a \textbf{novel head} attached to an SIOD model, and its training serves as the initial stage for the ODD-VOD framework. The primary goal of the ODD Predictor is to estimate the ODD score for a given input frame. 

Four steps are involved in training an ODD Predictor, as illustrated in Figure \ref{fig:odd_predictor}. First, an SIOD model is trained on the training dataset, or a pre-trained detector can be used. Second, the trained SIOD model is executed to produce detection results. Then, with the detection and ground-truth results collected, the method described in Section \ref{sec:odd_defi} is employed to generate the ODD score set, which serves as the supervision signal for ODD Predictor training. Finally, the ODD Predictor is trained with the backbone frozen, ensuring that the training process does not affect the object detection (DET) head.

The ODD Predictor is a lightweight auxiliary head with four layers: one layer of convolutional network with 3 by 3 kernel, one layer of adaptive average pooling with 7 by 7 kernel, and two fully connected layers. For a video frame $x_i$, the ODD score $y_i$ can be calculated as:
\begin{equation}
    y_i = ODD(f(x_i)),
\end{equation}
where $f$ is the detector's backbone and $x_i$ is the training image. $y_i$ ranges from 0 to 1. And we use the smooth L1 loss \cite{fast_rcnn} to optimize the ODD Predictor:
\begin{equation}
    ODDloss(z_i) = \left\{\begin{matrix}
    0.5 z_i^2, \textrm{if} \left| z_i \right| < 1 \\
     \\
    \left| z_i \right| - 0.5, \textrm{otherwise}
    \end{matrix}\right.
\end{equation}
Here, $z_i = gt_i - 10 y_i$. $gt_i$ is the ground truth ODD score of frame $x_i$, which can be obtained in step 3. Note that we magnify the value of $y_i$ by 10 times to increase the convergence speed when calculating the ODD loss.

\minisection{ODD Scheduler for Faster Detection (ODD-A)}

\label{sec:odd_scheduler}
Figure \ref{fig:odd_scheduler} illustrates the working process of the ODD Scheduler, which functions as a hybrid detection pipeline. It comprises two object detectors: a Still Image Object Detector (SIOD) and a Video Object Detector (VOD). While the SIOD can detect objects more rapidly but with reduced detection accuracy, the VOD can achieve superior detection results, albeit at a slower speed.

During the inference stage, the ODD Predictor first assigns scores to the frames. If the predicted ODD score for the current frame is below the ODD threshold (indicating an easy-to-detect frame), the SIOD detection head will directly process it. Otherwise, the ODD Scheduler will delegate this frame (considered hard-to-detect) to the Video Object Detector, representing the dispatching process. The ODD Scheduler subsequently collects all detected results in their original order.

Thus, the ODD-VOD framework processes a video in two rounds. In the first round, the SIOD equipped with the ODD Predictor evaluates the entire video and performs detection when necessary. In the second round, the VOD detects the remaining frames.

\begin{figure}[tbp]
    \centering
    \includegraphics[width=0.5\textwidth]{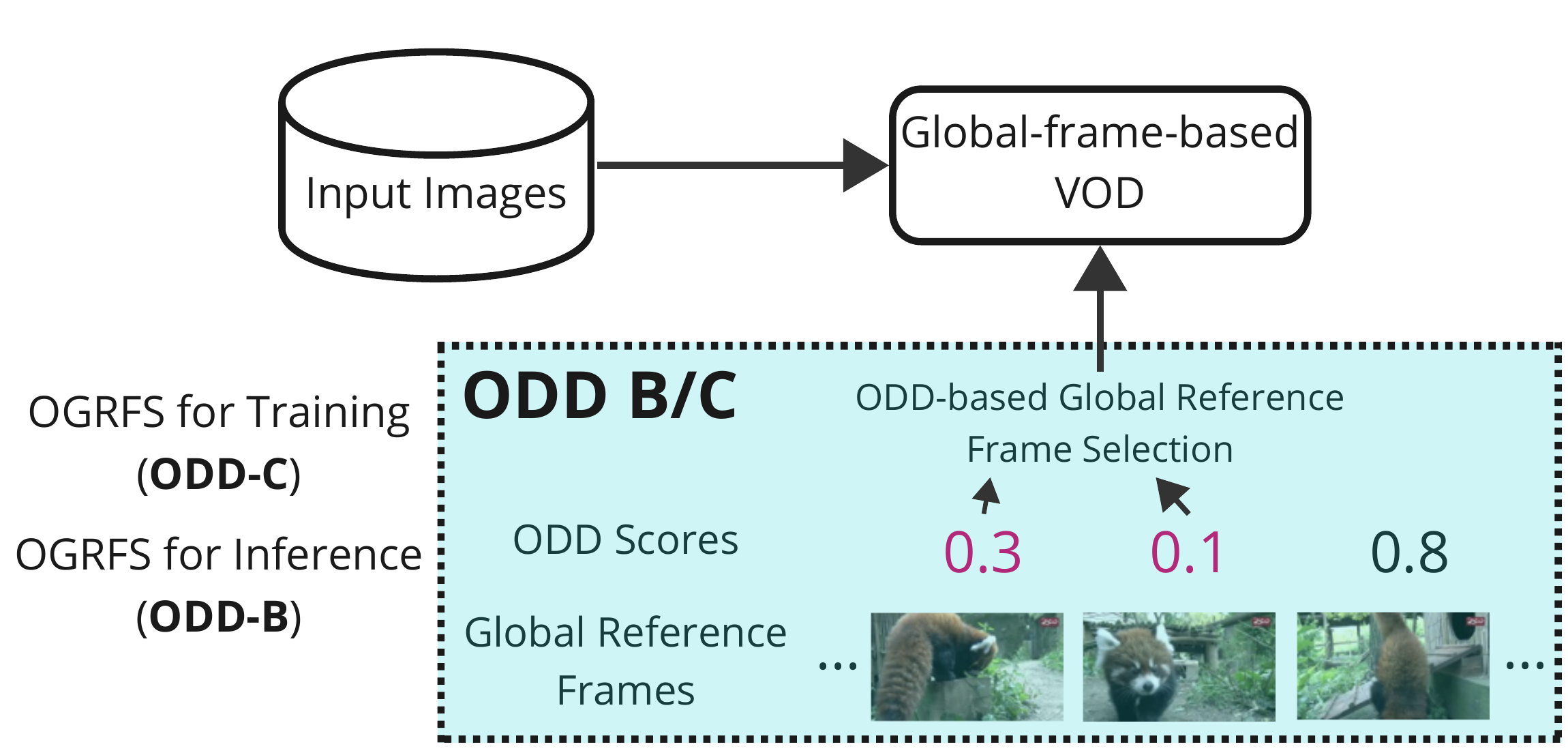}
    \caption{OGRFS for Better Detection. Frames with lower ODD scores tend to yield more accurate detection outcomes and can serve as reliable reference frame for other hard-to-detect frames during feature aggregation, thus improving overall performance.}
    \label{fig:ogrfs}
\end{figure}

\begin{table*}[tbp]
\centering
\caption{Main results for the full ODD-VOD framework (ODD-Full, i.e. ODD-A+[B+C], all ODD components used when applicable to the VOD model). We use ResNet-50 backbone for all models. The bold numbers mean settings on which acceleration is achieved without sacrificing accuracy, and the biggest lossless acceleration rate is shown in the rightmost column. On the other hand, when the ODD Threshold is set to near 0, we can get maximum accuracy improvement in this table, and the corresponding mAP gains over the original VOD models are given in the second rightmost column.}
\label{tab:odd_inference}
\resizebox{\textwidth}{!}
{
\begin{tabular}{c|c|ccccccccccccc|c|c}
\toprule[1pt]
\multirow{2}{*}{\textbf{VOD Model}} & \multirow{2}{*}{\textbf{\begin{tabular}[c]{@{}c@{}}ODD-\\ Strategy\end{tabular}}} & \multirow{2}{*}{\textbf{}} & \multicolumn{12}{c|}{\textbf{ODD Theshold}} & \multirow{2}{*}{\textbf{\begin{tabular}[c]{@{}c@{}}Original  \\ Results \end{tabular}}} & \multirow{2}{*}{\textbf{\begin{tabular}[c]{@{}c@{}}Lossless\\ Acc. Rate\end{tabular}}} \\
 &  &  & \textbf{0.9} & \textbf{0.8} & \textbf{0.7} & \textbf{0.6} & \textbf{0.5} & \textbf{0.4} & \textbf{0.3} & \textbf{0.25} & \textbf{0.2} & \textbf{0.15} & \textbf{0.1} & \textbf{0.05}  &  \\ \hline
\multirow{2}{*}{\begin{tabular}[c]{@{}c@{}}FGFA\\ \cite{fgfa}\end{tabular}} & \multirow{2}{*}{\begin{tabular}[c]{@{}c@{}}ODD-\\ A\end{tabular}} & \multicolumn{1}{c|}{mAP} & 72.1 & 72.8 & 73.1 & 73.4 & 73.8 & 71.2 & 74.5 & \textbf{74.7} & \textbf{74.7} & \textbf{74.7} & \textbf{74.7} & \textbf{74.7} & 74.7 & \multirow{2}{*}{$\uparrow$104.2\%} \\
 &  & \multicolumn{1}{c|}{FPS} & 29.6 & 27.5 & 25.9 & 23.8 & 19.4 & 17.2 & 15.3 & \textbf{14.5} & \textbf{13.6} & \textbf{13.1} & \textbf{12.2} & \textbf{11.4}  & 7.1 &  \\ \hline
\multirow{2}{*}{\begin{tabular}[c]{@{}c@{}}SELSA\\ \cite{selsa}\end{tabular}} & \multirow{2}{*}{\begin{tabular}[c]{@{}c@{}}ODD-\\ A+B+C\end{tabular}} & \multicolumn{1}{c|}{mAP} & 73.3 & 74.4 & 75.0 & 75.8 & 76.7 & 77.6 & \textbf{78.7} & \textbf{79.2} & \textbf{79.5} & \textbf{79.7} & \textbf{79.8} & \textbf{80.0}  & 78.4 & \multirow{2}{*}{$\uparrow$ 72.4\%} \\
 &  & \multicolumn{1}{c|}{FPS} & 34.1 & 33.7 & 30.7 & 30.3 & 25.6 & 22.7 & \textbf{20.0} & \textbf{19.1} & \textbf{18.1} & \textbf{16.6} & \textbf{15.4} & \textbf{14.2}  & 11.6 &  \\ \hline
\multirow{2}{*}{\begin{tabular}[c]{@{}c@{}}TROIA\\ \cite{temporal_roi_align}\end{tabular}} & \multirow{2}{*}{\begin{tabular}[c]{@{}c@{}}ODD-\\ A+B+C\end{tabular}} & \multicolumn{1}{c|}{mAP} & 73.9 & 75.5 & 76.3 & 76.7 & 77.7 & 78.8 & 79.7 & \textbf{79.9} & \textbf{80.1} & \textbf{80.2} & \textbf{80.4}  & \textbf{80.6} & 79.8 & \multirow{2}{*}{$\uparrow$ 66.7\%} \\
 &  & \multicolumn{1}{c|}{FPS} & 28.7 & 24.6 & 22.5 & 19.7 & 15.9 & 13.0 & 10.8 & \textbf{10.0} & \textbf{9.1} & \textbf{8.4} & \textbf{7.6} & \textbf{6.9}  & 6.0 &  \\ \hline
\multirow{2}{*}{\begin{tabular}[c]{@{}c@{}}RDN(local)\\ \cite{rdn}\end{tabular}} & \multirow{2}{*}{\begin{tabular}[c]{@{}c@{}}ODD-\\ A\end{tabular}} & \multicolumn{1}{c|}{mAP} & 72.0 & 72.5 & 72.7 & 73.0 & 73.7 & 74.2 & 74.8 & 75.1 & 75.3 & 75.5 & 75.6 & \textbf{75.7}  & 75.7 & \multirow{2}{*}{$\uparrow$ 13.9\%} \\
 &  & \multicolumn{1}{c|}{FPS} & 26.6 & 23.1 & 21.4 & 19.0 & 15.7 & 13.6 & 11.4 & 10.8 & 9.8 & 9.5 & 8.9 & \textbf{8.2} & 7.2 &  \\ \hline
\multirow{2}{*}{\begin{tabular}[c]{@{}c@{}}RDN(global)\\ \cite{mega}\end{tabular}} & \multirow{2}{*}{\begin{tabular}[c]{@{}c@{}}ODD-\\ A+B+C\end{tabular}} & \multicolumn{1}{c|}{mAP} & 73.1 & 74.1 & 74.9 & 75.6 & 76.5 & \textbf{77.6} & \textbf{78.9} & \textbf{79.2} & \textbf{79.4} & \textbf{79.7} & \textbf{79.9} & \textbf{80.1}  & 77.6 & \multirow{2}{*}{$\uparrow$ 70.3\%} \\
 &  & \multicolumn{1}{c|}{FPS} & 36.2 & 34.5 & 33.0 & 31.5 & 27.9 & \textbf{24.7} & \textbf{22.7} & \textbf{21.7} & \textbf{19.5} & \textbf{18.7} & \textbf{17.6} & \textbf{16.0}  &14.5 &  \\ \hline
\multirow{2}{*}{\begin{tabular}[c]{@{}c@{}}MEGA\\ \cite{mega}\end{tabular}} & \multirow{2}{*}{\begin{tabular}[c]{@{}c@{}}ODD-\\ A+B+C\end{tabular}} & \multicolumn{1}{c|}{mAP} & 72.3 & 73.0 & 73.4 & 74.2 & 75.2 & 76.1 & \textbf{77.2} & \textbf{77.6} & \textbf{77.9} & \textbf{78.2} & \textbf{78.4} & \textbf{78.5}  & 77.0 & \multirow{2}{*}{$\uparrow$ 112.5\%} \\
 &  & \multicolumn{1}{c|}{FPS} & 25.5 & 21.9 & 19.5 & 18.2 & 14.3 & 12.4 & \textbf{10.2} & \textbf{9.6} & \textbf{8.7} & \textbf{8.2} & \textbf{7.7} & \textbf{6.9}  & 4.8 &  \\ \bottomrule[1pt]
\end{tabular}
}
\end{table*}

\begin{table}[tbp]
\centering
\caption{Main results for ODD-VOD. We use ResNet-101 backbone for all models.}
\label{tab:r101_odd_inference}
\resizebox{0.5\textwidth}{!}
{
\begin{tabular}{cl|cccccc|c|c}
\toprule[1pt]
\multicolumn{2}{c|}{\multirow{2}{*}{\textbf{Model}}} & \multicolumn{6}{c|}{\textbf{ODD Threshold}} & \multirow{2}{*}{\textbf{\begin{tabular}[c]{@{}c@{}}Original\\ Results\end{tabular}}} & \multirow{2}{*}{\textbf{\begin{tabular}[c]{@{}c@{}}Acc. \\ Rate\end{tabular}}} \\
\multicolumn{2}{c|}{} & 0.8 & 0.4 & 0.2 & 0.15 & 0.1 & 0.05 &  &  \\ \hline
\multicolumn{1}{c|}{\multirow{2}{*}{FGFA}} & mAP & 76.9 & 77.1 & 77.6 & \textbf{77.8} & \textbf{77.8} & \textbf{77.8} & 77.8 & \multirow{2}{*}{ $\uparrow$ 98.4\%} \\
\multicolumn{1}{c|}{} & FPS & 25.7 & 16.7 & 12.5 & \textbf{12.1} & \textbf{11.1} & \textbf{10.6} & 6.3 &  \\ \hline
\multicolumn{1}{c|}{\multirow{2}{*}{SELSA}} & mAP & 78.4 & 80.2 & \textbf{81.7} & \textbf{82.1} & \textbf{82.5} & \textbf{82.7} & 81.5 & \multicolumn{1}{l}{\multirow{2}{*}{$\uparrow$ 50.5\%}} \\
\multicolumn{1}{c|}{} & FPS & 28.3 & 17.9 & \textbf{15.8} & \textbf{15.0} & \textbf{13.9} & \textbf{12.8} & 10.5 & \multicolumn{1}{l}{} \\ \hline
\multicolumn{1}{c|}{\multirow{2}{*}{TROIA}} & mAP & 80.6 & 81.5 & \textbf{82.7} & \textbf{82.9} & \textbf{83.6} & \textbf{83.9} & 82.6 & \multirow{2}{*}{$\uparrow$ 62.7\%} \\
\multicolumn{1}{c|}{} & FPS & 20.5 & 11.7 & \textbf{8.3} &  \textbf{7.5} & \textbf{6.8} & \textbf{6.2} & 5.1 &  \\ \hline
\multicolumn{1}{c|}{\multirow{2}{*}{\begin{tabular}[c]{@{}c@{}}RDN\\ (Local)\end{tabular}}} & mAP & 77.2 & 78.9 & 80.5 & 81.2 & 81.5 & \textbf{81.7} & 81.7 & \multirow{2}{*}{$\uparrow$ 18.0\%} \\
\multicolumn{1}{c|}{} & FPS & 21.6 & 12.3 & 8.9 & 8.3 & 7.8 & \textbf{7.2} & 6.1 &  \\ \hline
\multicolumn{1}{c|}{\multirow{2}{*}{\begin{tabular}[c]{@{}c@{}}RDN\\ (Global)\end{tabular}}} & mAP & 80.4 & 81.5 & \textbf{82.3} & \textbf{82.8} & \textbf{82.3} & \textbf{83.6} & 81.7 & \multirow{2}{*}{$\uparrow$ 45.6\%} \\
\multicolumn{1}{c|}{} & FPS & 29.6 & 18.2 & \textbf{16.6} & \textbf{16.1} & \textbf{15.7} & \textbf{15.2} & 11.4 &  \\ \hline
\multicolumn{1}{c|}{\multirow{2}{*}{MEGA}} & mAP & 80.8 & 81.7 & 82.8 & \textbf{83.2} & \textbf{83.9} & \textbf{84.1} & 82.9 & \multirow{2}{*}{$\uparrow$ 86.1\%} \\
\multicolumn{1}{c|}{} & FPS & 19.6 & 10.9 & 7.4 & \textbf{6.7} & \textbf{5.9} & \textbf{5.3} & 3.6 &  \\ 
\bottomrule[1pt]
\end{tabular}
}
\end{table}

\minisection{OGRFS for Better Detection (ODD-B/C)}
 
\label{sec:ogrfs}
For global-frame-based VODs, selecting global frames from the entire video as reference frames for aggregation is essential. However, existing sampling strategies are typically naive. The most common strategy involves randomly selecting several frames as global frames. In ODD-VOD, we propose the ODD-based Global Reference Frame Selection (OGRFS) method as a more sophisticated selection strategy, as illustrated in Figure \ref{fig:ogrfs}. Given a number $k$ as a parameter for the reference frames to be aggregated, OGRFS selects the frames with the $k$ lowest ODD scores to create a global frame pool.

OGRFS can be employed in two distinct ways. One approach is to directly incorporate OGRFS into the VOD inference stage (ODD-B), which enhances detection accuracy without re-training the models. Alternatively, OGRFS can be further utilized to train global-frame-based VOD models (ODD-C), resulting in superior detection accuracy compared to the inference-only version. In summary, OGRFS helps avoid selecting deteriorated frames as global reference frames during both training and inference stages.

\section{Experiments}
\subsection{Experiment Setup}

\minisection{Dataset and Evaluation.}We conduct our experiments on the ImageNet VID dataset \cite{imagenet}, which is the most widely used benchmark for video object detection \cite{fgfa, mega, selsa}. The dataset consists of 3,862 video snippets for training and 555 video snippets for validation. All video frames are fully annotated with bounding boxes and object categories, covering over 30 categories.

To evaluate the performance of detection results, we follow the common practice in VOD \cite{transvod_lite, yolov} and report mean Average Precision (mAP) at 0.50 IoU on the validation set as the accuracy metric and runtime as the speed metric. It is worth noting that runtime may vary due to system factors. Therefore, we typically take the average of multiple measurements. Moreover, two types of runtime are widely used in previous work: runtime (ms) and runtime (FPS). We choose runtime (FPS) because we believe that FPS provides a more intuitive understanding of a model's inference speed.

\minisection{Training and Inference Details.} We implemented our code using PyTorch 1.12.0. For the main experiments, we employed the MMTracking toolbox \cite{mmtrack}, a widely-used open-source platform for video object detection, to implement the methods and compare their performance. We utilized Faster R-CNN with ResNet-50 and ResNet-101 \cite{resnet} backbones, initialized with ImageNet pre-trained weights, as the still image object detector. In the generalizability experiments (Section \ref{sec:generalisability}), we implemented our ODD-VOD framework using the corresponding official codebases \cite{transvod_lite,yolov}. It is important to note that image preprocessing varies across different methods. For YOLOV experiments, we resized input frames to $576 \times 576$. In contrast, for other experiments, we resized input frames to $1000 \times 600$.

For training, we employed three NVIDIA GeForce RTX 3090 GPUs to train the ODD predictor, Faster R-CNN, and the VOD models with OGRFS (ODD-B) when applicable, in parallel. The process unfolded as follows: first, we trained Faster R-CNN for 7 epochs and then generated the ODD ground truth scores on the training dataset, with the near-positive threshold ($t_1$) set to 0.3 and the positive threshold ($t_2$) set to 0.5. After generating the ODD ground truth scores, we froze the backbone to train the ODD Predictor. We set the training iteration to 220k. The initial learning rate was set to $3.75 \times 10^{-3}$ and was divided by 10 at 110k and 165k iterations, respectively. Subsequently, we trained several VOD models with OGRFS. The training iteration remained the same as for the ODD Predictor. We followed the dataset protocols widely used in \cite{fgfa,mega, transvod, yolov}.

In the main experiments, we evaluated ODD-VOD's performance on six different VODs to test its effectiveness: FGFA \cite{fgfa}, SELSA \cite{selsa}, MEGA \cite{mega}, RDN(local) \cite{rdn}, RDN(global), and Temporal RoI Align (TROIA) \cite{temporal_roi_align}. To obtain comparable results for inference speed, we utilized a single NVIDIA GeForce RTX 3090 GPU to evaluate the ImageNet VID dataset and set the batch size in Faster R-CNN and VOD to 1. In the generalizability experiments, we directly reused ODD predicted scores obtained from the main experiments and combined ODD-VOD with the two latest state-of-the-art models, namely YOLOV (YOLO series) \cite{yolov} and TransVOD\_Lite (DETR series) \cite{transvod_lite}, to achieve better performance. The batchsize of SIODs was aligned with the original models, i.e., 1 for SELSA, 12 for TransVOD\_Lite and 32 for YOLOV.

\subsection{Main Results}

Tables \ref{tab:odd_inference} and \ref{tab:r101_odd_inference} present the detection accuracy and speed on various VOD models with different ODD thresholds, employing two different backbones (ResNet-50 and ResNet-101). The ODD strategy in Table \ref{tab:odd_inference} indicates the ODD components (ODD-A, ODD-B, and ODD-C) included in the models. In Table \ref{tab:r101_odd_inference}, the corresponding ODD strategies remain consistent. The right column displays the results of the original VOD model. Among these VOD models, FGFA and RDN (local) are local-frame-based methods; SELSA, TROIA, and RDN (global) are global-frame-based; MEGA is both-frame-based. These six models represent all types of typical precision-oriented VODs. The bold numbers in the tables signify the accuracy lossless ODD settings, which are recommended speed-accuracy trade-offs.

The ODD threshold is used by ODD-A for frame assignment to either an SIOD or a VOD process. If the predicted ODD score falls below the ODD threshold, Faster R-CNN will detect the frame. Otherwise, the frame will be processed by the VOD. Consequently, as the ODD threshold decreases, detection accuracy increases while detection speed slows down. The ODD threshold can be flexibly adjusted to cater to different application requirements. Furthermore, we find that setting the ODD Threshold to 0.2 allows most SIOD-VOD combinations to achieve lossless acceleration. Therefore, \textbf{0.2  is a recommended ODD threshold} for our ODD-VOD framework.

\begin{table}[tp]
\centering
\caption{Generalisability for ODD Metric in Different ODD-VOD Combinations}
\label{tab:generality}
\resizebox{0.5\textwidth}{!}
{
\begin{tabular}{l|l|cccccc|c}
\toprule[1pt]
\multicolumn{1}{c|}{\multirow{2}{*}{\textbf{\begin{tabular}[c]{@{}c@{}}Still Image\\ Detector\end{tabular}}}} & \multicolumn{1}{c|}{\multirow{2}{*}{\textbf{\begin{tabular}[c]{@{}c@{}}Video Object\\ Detector\end{tabular}}}} & \textbf{} & \multicolumn{5}{c|}{\textbf{ODD Threshold}} & \multirow{2}{*}{\textbf{\begin{tabular}[c]{@{}c@{}}Original\\ Results\end{tabular}}} \\
\multicolumn{1}{c|}{} & \multicolumn{1}{c|}{} &  & 0.7 & 0.4 & 0.2 & 0.1 & 0.05 &  \\ \hline
\multirow{2}{*}{\begin{tabular}[c]{@{}l@{}}Deformable DETR\\ (Swin-Base)\end{tabular}} & \multirow{2}{*}{\begin{tabular}[c]{@{}l@{}}TransVOD Lite\\ (Swin-Base)\end{tabular}} & \multicolumn{1}{c|}{mAP} & 88.3 & 89.1 & 89.9 & 90.1 & 90.1 & 90.1 \\
 &  & \multicolumn{1}{c|}{FPS} & 16.1 & 14.7 & 13.0 & 11.7 & 11.0 & 9.9 \\ \hline
\multirow{2}{*}{YOLOX-s} & \multirow{2}{*}{YOLOV-s} & \multicolumn{1}{c|}{mAP} & 73.0 & 74.8 & 76.3 & 77.5 & 77.9 & 77.9 \\
 &  & \multicolumn{1}{c|}{FPS} & 193.8 & 184.0 & 176.9 & 170.3 & 166.1 & 161.3 \\ \hline
\multirow{2}{*}{YOLOX-l} & \multirow{2}{*}{YOLOV-l} & \multicolumn{1}{c|}{mAP} & 77.7 & 79.2 & 81.2 & 82.7 & 83.4 & 83.4 \\
 &  & \multicolumn{1}{c|}{FPS} & 142.0 & 135.5 & 125.9 & 120.1 & 117.3 & 113.0 \\ \hline
\multirow{2}{*}{YOLOX-x} & \multirow{2}{*}{YOLOV-x} & \multicolumn{1}{c|}{mAP} & 83.1 & 84.3 & 85.2 & 85.6 & 85.6 & 85.6 \\
 &  & \multicolumn{1}{c|}{FPS} & 105.7 & 95.5 & 90.1 & 84.4 & 82.5 & 78.1 \\ \hline
\multirow{2}{*}{YOLOX-s} & \multirow{2}{*}{YOLOV-x} & \multicolumn{1}{c|}{mAP} & 74.5 & 79.2 & 82.4 & 83.6 & 84.3 & 85.6 \\
 &  & \multicolumn{1}{c|}{FPS} & 182.6 & 151.4 & 128.0 & 103.7 & 91.1 & 78.1 \\
 \bottomrule[1pt]
\end{tabular}
}
\end{table}

Furthermore, Table \ref{tab:odd_inference} validates the existence of the \textit{over-aggregation} phenomenon. When we set the ODD threshold to a specific number or below, we can achieve the same detection results with a faster inference speed. For instance, in FGFA detection, when we set the ODD threshold to 0.25 for ResNet-50 and 0.15 for ResNet-101, ODD-VOD can correspondingly achieve the same mAP as the initial results with 104\% and 98.4\% inference speed, respectively. For global-frame-based and both-frame-based methods (SELSA, TROIA, RDN (global), and MEGA), OGRFS makes the aggregation operation more efficient, resulting in better detection accuracy and faster detection outcomes. For example, when we set the ODD threshold to 0.3, SELSA can achieve a 79.2 mAP and 20.0 FPS (compared to the initial results: 78.4 mAP and 11.6 FPS).

\subsection{Generalisability for ODD-VOD (ODD-A).} 
\label{sec:generalisability}

In the main experiments, we use Faster R-CNN as the SIOD to verify the effectiveness of ODD-A with the defined ODD metric in different VODs. We then further explore the generalizability of ODD-VOD on various combinations of SIODs and VODs. Here we directly reuse the predicted ODD scores mentioned earlier. We include Deformable DETR \cite{deformable_detr} and YOLOX \cite{yolox} as alternative SIOD options to Faster R-CNN and test both SIODs on the very recent and competitive VOD models, TransVOD Lite \cite{transvod_lite} and YOLOV \cite{yolov}, to validate the generalizability and flexibility of ODD-VOD. Table \ref{tab:generality} displays the detection results at different ODD thresholds. We observe consistent performance gains for both SIODs combined with TransVOD Lite and YOLOV. This confirms that the ODD-A component in ODD-VOD is robust to the choice of SIOD. Therefore, we argue that the proposed ODD metric is suitable for various detectors (detector-agnostic) because most detectors share common challenging detection cases, such as small and blurry objects.

\begin{table}[tbp]
\centering
\caption{OGRFS for Global-based VOD Training \& Inference.}
\label{tab:ogrfs}
\resizebox{0.3\textwidth}{!}
{
\begin{tabular}{l|lccc}
\toprule[1pt]
\multicolumn{1}{c|}{\multirow{3}{*}{\textbf{Model}}} & \multicolumn{1}{c}{ODD-B} & \textbf{} & \ding{52} & \ding{52} \\
\multicolumn{1}{c|}{} & \multicolumn{1}{c}{ODD-C} &  &  & \ding{52} \\ \cline{2-5} 
\multicolumn{1}{c|}{} & \multicolumn{1}{c|}{\textbf{Backbone}} & \multicolumn{3}{c}{\textbf{mAP@0.50}} \\ \hline
\multirow{2}{*}{\begin{tabular}[c]{@{}l@{}}RDN(Global)\\ \cite{mega}\end{tabular}} & \multicolumn{1}{l|}{ResNet-50} & 77.6 & 79.1 & 80.1 \\
 & \multicolumn{1}{l|}{ResNet-101} & 82.5 & 83.0 & 83.6 \\ \hline
\multirow{2}{*}{\begin{tabular}[c]{@{}l@{}}TROIA\\ \cite{temporal_roi_align}\end{tabular}} & \multicolumn{1}{l|}{ResNet-50} & 79.8 & 80.1 & 80.6 \\
 & \multicolumn{1}{l|}{ResNet-101} & 82.6 & 83.4 & 84.0 \\ \hline
\multirow{2}{*}{\begin{tabular}[c]{@{}l@{}}SELSA\\ \cite{selsa}\end{tabular}} & \multicolumn{1}{l|}{ResNet-50} & 78.4 & 79.7 & 80.1 \\
 & \multicolumn{1}{l|}{ResNet-101} & 81.5 & 81.8 & 82.7 \\ \bottomrule[1pt]
\end{tabular}
}
\end{table}

\subsection{Evaluation of OGRFS (ODD-B/C)}

Table \ref{tab:ogrfs} shows the effects of ODD for global-frame-based VOD training and inference. The results show that the OGRFS can improve detection accuracy even with only a pre-trained model in the inference stage (ODD-B). We use SELSA and TROIA (78.4 and 79.8 mAP correspondingly) with the model and weights directly downloaded  from the Open-MMLab \footnote{https://mmtracking.readthedocs.io/en/latest/model\_zoo.html}. OGRFS achieves better detection accuracy (79.7 and 80.1 mAP correspondingly) without retraining. This means ODD-B can be a low-cost plug-in for better accuracy, for any pretrained VOD model in this category. Furthermore, if we use OGRFS for both retraining and inference (ODD-B+C), ODD-VOD outperforms all existing VOD models consistently.


\subsection{Design of the ODD Quantifying Metric.} 

In Section \ref{sec:odd_defi}, we defined the categorized output bounding boxes into four kinds of results: positive, negative, near-positive, and multi-positive. The notion of positive and negative results is widely used in F1-score and mAP. In this ablation study, we prove the effectiveness of near-positive and multi-positive results. Table \ref{tab:ab_study} shows the detection results (accuracy and speed) of ODD-based SELSA with different components with the ODD threshold from 0.7, 0.4 to 0.2. When we use all components, the detection results can perform best in most cases. If we use two of them, the performance can be better than only using positive \& negative detections.

\begin{table}[htbp]
\caption{Ablation Study for Object Detection Difficulty Design}
\label{tab:ab_study}
\centering
\resizebox{0.35\textwidth}{!}
{
\begin{tabular}{c|cc|ccc}
\toprule[1pt]
\multirow{3}{*}{Component} & \multicolumn{2}{c|}{positive\&negative} & \ding{52} & \ding{52} & \ding{52} \\
 & \multicolumn{2}{c|}{near-positive} &  & \ding{52} & \ding{52} \\
 & \multicolumn{2}{c|}{multi-positive} &  &  & \ding{52} \\ \hline
\multirow{6}{*}{\begin{tabular}[c]{@{}c@{}}ODD\\ Threshold\end{tabular}} & \multirow{2}{*}{0.7} & mAP & 74.8 & 75.0 & \textbf{75.1} \\
 &  & FPS & 30.5 & \textbf{31.4} & 30.7 \\ \cline{2-6} 
 & \multirow{2}{*}{0.4} & mAP & 77.1 & 77.3 & \textbf{77.7} \\
 &  & FPS & 21.9 & 19.6 & \textbf{22.7} \\ \cline{2-6} 
 & \multirow{2}{*}{0.2} & mAP & 79.0 & \textbf{79.3} & 79.2 \\
 &  & FPS & 17 & 14.5 & \textbf{18.1} \\  \bottomrule[1pt]
\end{tabular}
}
\end{table}

\subsection{Comparison with Efficient VOD Methods}

\minisection{Comparison with Plug-in Efficient VODs.}

Plug-in efficient VOD methods can be attached to other video object detectors for faster detection speed without significant accuracy losses. The proposed ODD-VOD framework addresses the issue of over-aggregation by alternating between the SIOD and VOD, which is a typical plug-in approach. However, only a few methods are designed in plug-in form. To the best of our knowledge, DFA \cite{dfa} is also a plug-in efficient method addressing this challenge by designing a plug-in module to dynamically aggregate features for off-the-shelf video object detectors. It can reduce the computational cost and improve the detection speed when attached to other VOD models. It is worth mentioning that some plug-in VOD methods \cite{tfblender,boxmask} also exist which enhance feature aggregation for better detection accuracy. However, these methods are not considered efficient approaches because they sacrifice detection speed to achieve higher precision. In Table \ref{tab:plugin_compare}, we compare ODD-VOD with DFA (including Vanilla DA and Deformable DA) to show the detection results changes, including FPS and mAP. Since DFA is deployed on a different hardware platform, one reasonable way is to compare the gains of the base VOD models after the enhancement frameworks are applied. Overall, ODD-enhanced methods can achieve the best results without accuracy losses, while DFA will lead to a decline in detection accuracy. This indicates that DFA can partially alleviate the over-aggregation problem for speedups, but it comes at the cost of the model's accuracy.

\begin{table}[tbp]
\caption{Comparison with plug-in efficient VODs. Since the results of Vanilla DA and Deformable DA are reported from different platforms,  we compare relative gains here}
\label{tab:plugin_compare}
\centering
\resizebox{0.5\textwidth}{!}
{
\begin{tabular}{llll}
\toprule[1pt]
Methods & \textbf{Backbone} & \textbf{FPS Gains} & \textbf{mAP Gains} \\ \hline
FGFA + Vanilla DA & ResNet-50 &$\uparrow 2.1_{~ 5.8\rightarrow7.9}$ & $\downarrow0.4_{~74.3\rightarrow73.9}$  \\
FGFA + Deformable DA & ResNet-50 & $\uparrow1.8_{~5.8\rightarrow7.6}$ & $\downarrow0.2_{~74.3\rightarrow74.1}$ \\
FGFA + ODD (ours) & ResNet-50 & 
$\mathbf{\uparrow7.4_{~7.1\rightarrow14.5}}$  & 
$\mathbf{0.0_{~74.7\rightarrow74.7}}$ \\ \hline
SELSA + Vanilla DA & ResNet-50 & $\uparrow4.4_{~5.0\rightarrow9.4}$ & $\downarrow1.4_{~77.9\rightarrow76.5}$ \\
SELSA + Deformable DA & ResNet-50 & $\uparrow3.8_{~5.0\rightarrow8.8}$ & $\downarrow0.4_{~77.9\rightarrow77.5}$ \\
SELSA + ODD (ours) & ResNet-50 & $\mathbf{\uparrow8.4_{~11.6\rightarrow20.0}}$ & $\mathbf{\uparrow0.3_{~78.4\rightarrow78.7}}$ \\ \hline
TROIA + Vanilla DA & ResNet-50 & $\uparrow2.4_{~1.5\rightarrow3.9}$ & $\downarrow1.2_{~79.0\rightarrow77.8}$ \\
TROIA + Deformable DA & ResNet-50 & $\uparrow2.1_{~1.5\rightarrow3.6}$ & $\downarrow0.2_{~79.0\rightarrow78.8}$ \\
TROIA + ODD (ours) & ResNet-50 & $\mathbf{\uparrow4.0_{~6.0\rightarrow10.0}}$ & $\mathbf{\uparrow0.1_{~79.8\rightarrow79.9}}$ \\ \hline
FGFA + Vanilla DA & ResNet-101 & $\uparrow2.4_{~5.1\rightarrow7.5}$ & $\downarrow0.4_{~77.6\rightarrow77.2}$  \\
FGFA + Deformable DA & ResNet-101 & $\uparrow2.0_{~5.1\rightarrow7.1}$ & $\downarrow0.1_{~77.6\rightarrow77.5}$ \\
FGFA + ODD (ours) & ResNet-101 & $\mathbf{\uparrow{5.8_{~6.3\rightarrow12.1}}}$ & $\mathbf{0.0_{~77.8\rightarrow77.8}}$ \\ \hline
SELSA + Vanilla DA & ResNet-101 & $\uparrow3.0_{~4.5\rightarrow7.5}$ & $\downarrow1.3_{~81.3\rightarrow80.0}$ \\
SELSA + Deformable DA & ResNet-101 & $\uparrow2.6_{~4.5\rightarrow7.1}$ & $\downarrow0.3_{~81.3\rightarrow81.0}$ \\
SELSA + ODD (ours) & ResNet-101 & $\mathbf{\uparrow4.5_{~10.5\rightarrow15.0}}$ & $ \mathbf{\uparrow0.6_{~81.5\rightarrow82.1}}$ \\ \hline
TROIA + Vanilla DA & ResNet-101 & $\mathbf{\uparrow2.4_{~1.2\rightarrow3.6}}$ & $\downarrow0.6_{~82.4\rightarrow81.8}$ \\
TROIA + Deformable DA & ResNet-101 & $\uparrow2.1_{~1.2\rightarrow3.3}$ & $ \downarrow0.4_{~82.4\rightarrow82.0}$ \\
TROIA + ODD (ours) & ResNet-101 & $ \mathbf{\uparrow2.4_{~5.1\rightarrow7.5}}$ & $\mathbf{ \uparrow0.3_{~82.6\rightarrow82.9}}$ \\ 
\bottomrule[1pt]
\end{tabular}
}
\end{table}

\begin{table}[htbp]
\centering
\caption{Comparison with SOTA VOD Methods}
\label{tab:comparision}
\resizebox{0.5\textwidth}{!}
{
\begin{tabular}{llcc}
\toprule[1pt]
\multicolumn{1}{c}{\textbf{Model}} & \multicolumn{1}{c}{\textbf{Backbone}} & \textbf{FPS} & \textbf{mAP@0.5} \\ \hline
\multicolumn{4}{c}{\textbf{Precision-Oriented VOD}} \\ 
\hline
SELSA \cite{selsa} & ResNet-50 & 11.6 & 78.4 \\
FGFA \cite{fgfa} & ResNet-101 & 6.3 & 77.8 \\
SELSA \cite{selsa} & ResNet-101 & 10.5 & 81.5 \\
TROIA \cite{temporal_roi_align} & ResNet-101 & 5.1 & 82.6 \\
LWDN \cite{lwdn} & ResNet-101 & 20(X) & 76.3 \\
MAMBA \cite{mamba} & ResNet-101 & 11.1(RTX) & 80.8 \\
MINet \cite{minet} & ResNet-101 & 7.5(V) & 80.2 \\
LRTR \cite{lrtr} & ResNet-101 & 10(X) & 80.6 \\
DSFNet \cite{dsfnet} & ResNet-101 & - & 84.1 \\
EBFA \cite{ebfa} & ResNet-101 & - & 84.8 \\
TransVOD Lite \cite{transvod,transvod_lite} & SwinBase & 9.9 & \textbf{90.1} \\
\hline
\multicolumn{4}{c}{\textbf{Unified Efficient VOD}} \\ \hline
SparseVOD \cite{sparse_vod} & ResNet-50 & 14.4(A100) & 80.3 \\
DFF \cite{dff} & ResNet-101 & 39.8(V100) & 73.5 \\
OGEMN \cite{ogemn} & ResNet-101 & 14.9(1080Ti) & 76.8 \\
QueryProp \cite{query_prop} & ResNet-101 & 26.8(X) & 82.3 \\
THP \cite{Zhu2017TowardsHP} & ResNet-101 & 13.0(K40) & 78.6 \\
DorT \cite{d_or_t} & ResNet-101 & 7.8(X) & 75.8 \\
PSLA \cite{psla} & ResNet-101 & 30.8(V)/18.7(X) & 77.1 \\
LSTS \cite{lsts} & ResNet-101 & 23.0(V) & 77.2 \\
EOVOD \cite{eovod} & YOLOX-m & 50.5(V100) & 74.5 \\
YOLOV \cite{yolov} & YOLOX-s & \textbf{161.3} & 77.9 \\
YOLOV \cite{yolov} & YOLOX-x & 78.1 & 85.6 \\ \hline
\multicolumn{4}{c}{\textbf{ODD-VOD (ours)}} \\ \hline
SELSA+ODD & ResNet-50 & 20.0 & 78.7 \\
FGFA+ODD & ResNet-101 & 12.1 & 77.8 \\
SELSA+ODD & ResNet-101 & 15.8 & 81.7 \\
TROIA+ODD & ResNet-101 & 6.6 & 82.7 \\
TransVOD Lite+ODD & SwinBase & 11.7 & \textbf{90.1} \\
YOLOV+ODD & YOLOX-s & \textbf{166.1} & 77.9 \\
YOLOV+ODD & YOLOX-x & 84.4 & 85.6 \\
\bottomrule[1pt]
\end{tabular}
}
\end{table}

\minisection{Comparison with Unified Efficient VOD Methods.} 

Unified methods deeply optimize the process of feature aggregation to obtain better detection speed. In Table \ref{tab:comparision}, we compare our ODD (Plug-in trade-off) with some representative state-of-the-art methods including some latest methods, e.g. \cite{query_prop, sparse_vod, yolov}. The remarks in column FPS mean different GPU platforms. Table \ref{tab:comparision} has three parts. The top part lists some precision-oriented VOD models which are designed to achieve better detection accuracy but may encounter the over-aggregation problem. The middle part displays some unified efficient VOD methods focusing on detection speed. The bottom part shows our methods. Finally, our ODD-VOD is competitive with all these methods and can achieve state-of-the-art performance when integrated with the latest methods, such as TransVOD Lite+ODD and YOLOV+ODD.

\begin{table}[htbp]
\caption{Proportion of Frames Processed by SIOD Detection Head.}
\label{tab:proportion}
    \centering
    \resizebox{0.48\textwidth}{!}
    {
    \begin{tabular}{ccccccc}
    \toprule[1pt]
    ODD Thresh & 0.9 & 0.7 & 0.5 & 0.3 & 0.2 & 0.1 \\ \hline
    Proportion & 90.8\% & 84.0\% & 71.5\% & 52.2\% & 39.6\% & 25.0\% \\ 
    \bottomrule[1pt]
    \end{tabular}
    }
\end{table}
\vspace{-0.3cm}

\subsection{Time Efficiency}

In comparison to other efficient VOD approaches, ODD-VOD is able to accelerate the detection speed directly from the base SIOD models. Consequently, we investigate the percentage of frames processed by the SIOD detection head under various ODD threshold settings (refer to Table \ref{tab:proportion}). It is evident that the ODD metric tends to assign higher scores to input frames. Remarkably, even when the ODD threshold is set to 0.1, approximately one-quarter of the frames are directly processed by the SIOD detection head, resulting in favorable speed-accuracy trade-offs for the majority of VOD methods.

We further compute the GFLOPs for each component of our ODD-VOD method. The GFLOPs of Faster R-CNN (r50) equipped with an ODD predictor were 131.63, with the detection head contributing 45.09 and the ODD predictor (head) contributing a mere 0.13. Consequently, the ODD head accounted for only 0.1\% of the model complexity, exerting negligible impact on detection speed. The VODs employed in our primary experiment were considerably more complex than SIOD; for example, SELSA had a GFLOP of 324.2. In summary, the ODD predictor can be regarded as a \textit{cost-effective} means to bridge the gap between SIOD and VOD.

\begin{figure}[htbp]
    \includegraphics[width=0.5\textwidth]{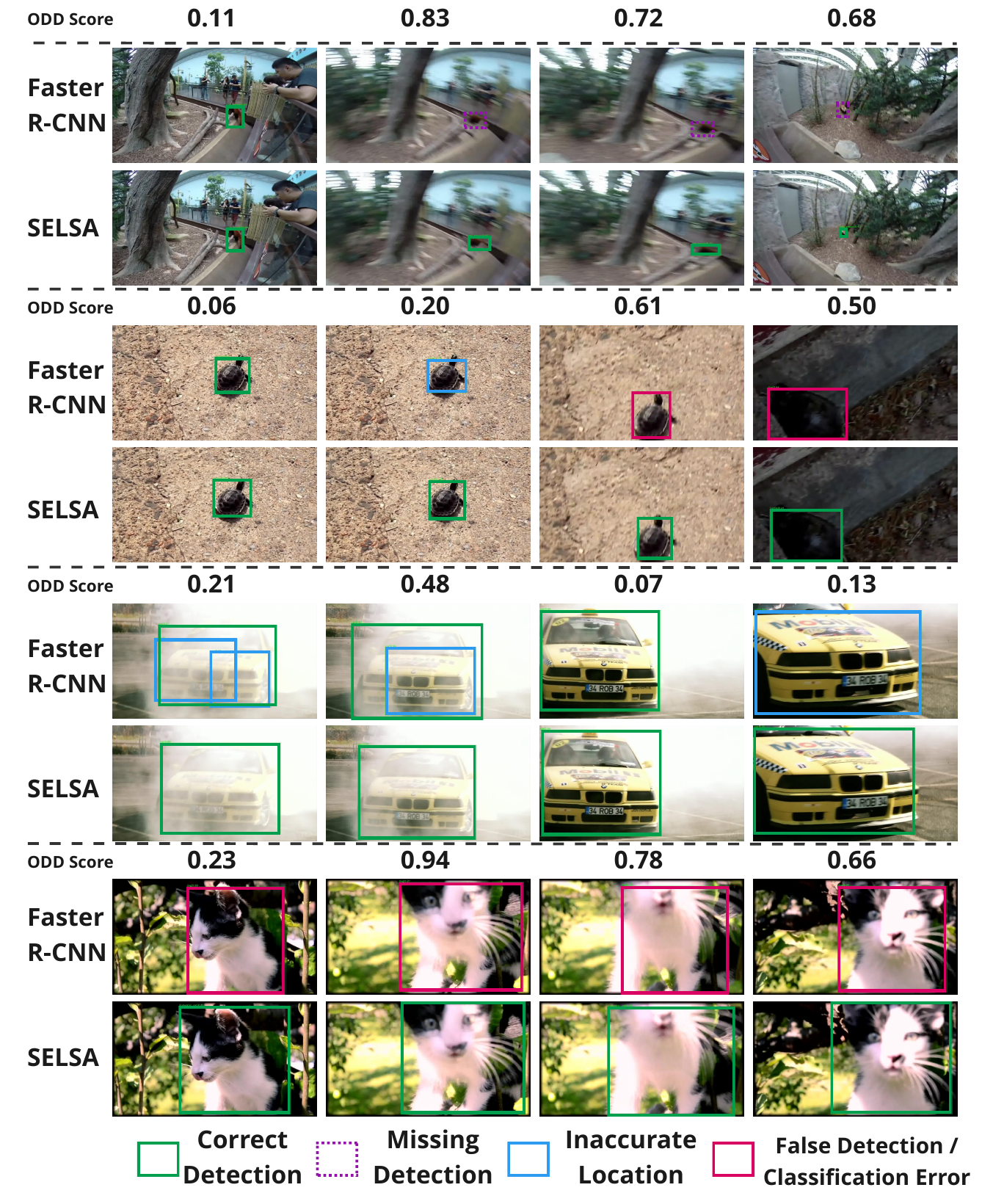}
    \caption{Visualization of the Relationship between Predicted ODD Scores and Detection Results.}
    \label{fig:vis}
\end{figure}

\subsection{Visualization}
\label{sec:visualization}

Figure \ref{fig:vis} shows more visualization results on the relationship between predicted ODD scores and the results for two different detectors (Faster R-CNN for a typical SIOD  and SELSA for a typical VOD). Predicted ODD score could accurately characterize the detection difficulty for Faster R-CNN in most cases. When the predicted ODD score is below 0.2, Faster R-CNN mainly outputs correct proposals. This phenomenon can also be proved by Table \ref{tab:odd_inference}: when we set the ODD threshold to 0.2, our ODD-VOD (ODD-Full) can achieve the same or better detection accuracy and faster detection speed compared with the original video object detector.
When the predicted ODD score is 0.7 or above, these frames will likely be challenging for Faster R-CNN to detect. Therefore, the feature aggregation module can get maximum detection benefit for the frames with ODD scores of 0.7 or above.

Furthermore, as an interpretation of what the ODD score is really measuring, we find that many frames with higher ODD scores exhibit motion blur, video defocus, part occlusion, rare poses, poor illumination, small object scales, and other attributes that affect detection difficulty. A high ODD score could be derived from a single or a combination of such attributes. Therefore, we argue that there is no simple heuristic/rule-based method to measure detection difficulty. Instead, the proposed ODD Predictor manages to capture that complex signal with the proper supervision from the ODD Metric, and is able to produce robust ODD predictions across different scenarios.

\section{Conclusion}
In this paper, we define an image-level metric named Object Detection Difficulty (ODD) to measure the difficulty of object detection for a given image. We also design an ODD-VOD framework using ODD scores to suppress over-aggregation for faster video object detection speed. In addition, using ODD scores, a module in ODD-VOD called OGRFS helps the global-frame-based VOD models select better reference frames for better detection accuracy. Finally, extensive experiments demonstrate that our method can achieve faster and better video object detection than other VOD methods tested.

\begin{acks}
This work was partially supported by CSIRO's Science Leader project R-91559.
\end{acks}

\clearpage

\bibliographystyle{ACM-Reference-Format}
\balance
\bibliography{sample-sigconf}

\clearpage

\appendix

\section{Relation With Image Quality Assessment}

Another existing work similar to our proposed ODD is called Image Quality Assessment (IQA) \cite{iqa, Ding2020ImageQA, Cheon2021PerceptualIQ}, which is widely used to assess image quality. However, ODD has enormous differences from IQA essentially. First, ODD pays attention to the objects in images, while IQA evaluates both the foreground and background in images, including the sky and clouds in images. Second, the influences on detection accuracy are not only about the quality of frames, other factors like object scales, and rare poses can also make detection difficult. The ODD score can comprehensively quantitatively measure all these factors (See Section \ref{sec:visualization}). In addition, the definition of IQA is more subjective, relying on humans' judgment, while ODD is defined from the perspective of an SIOD.

\section{Object Detection Difficulty Metric: the Algorithm Description}

\begin{algorithm}[htb]
    \caption{Calculate the ODD ground truth score for an image}
    \label{alg:odd}
\begin{flushleft}
    \textbf{Input}:  PredBbox, PredLabel, GtBbox, GtLabel \\
    \textbf{Parameter}: Threshold: near-positive ($t_1$), positive ($t_2$)\\
    \textbf{Output}: $ODD$
\end{flushleft}
    \begin{algorithmic}[1] 
        \STATE {PosList, NearList, MultiList, NegList $:=$ []} 
        \FOR{$l$ in \textbf{CONCAT}(PredLabel, GtLable)}
            \STATE {PredBboxL, GtBboxL $:=$ \textbf{GetBboxWithLabel}($l$)}
            \STATE {PredScoreL $:=$ \textbf{GetScoreWithLabel($l$)}}
            \STATE $IoU_l$ := \textbf{CalculateIoU}(PredBboxL, GtBboxL)
            \FOR{i in [0 ... NUM(PredBboxL)]}
                \STATE Weight := \textbf{PredScoreL(i)}
                \IF{$IoU_{li}$ is $max(IoU_l)$ \AND $IoU_{li} \geq t_2$}
                    \STATE APPEND(PosList, Weight)
                \ELSIF{$IoU_{li} \geq t_2$}
                    \STATE APPEND(MultiList, Weight)
                \ELSIF{$t_1 \leq IoU_{li} \leq t_2$}
                    \STATE APPEND(NearList, Weight)
                \ELSE
                    \STATE APPEND(NegList, Weight)
                \ENDIF
            \ENDFOR
        \ENDFOR
        \STATE $wp$, $wr$ := \textbf{CalPR}(PosList, NearList,MultiList, NegList)
        \STATE $ODD$ := $1 - 2 \cdot \frac{wp \times wr}{wp + wr + \varepsilon}$
        \STATE \textbf{return} $ODD$
    \end{algorithmic}
\end{algorithm}

In Section 3.1, we formally define the concept of Object Detection Difficulty (ODD) metric with equations (1 to 4). Equation 1 is used to divide the detection results of an SIOD into four categories and compute the $wp$. The form of the other three equations is very similar to that of the F1-Score.  

The procedure for calculating the ODD ground truth score is described in Algorithm \ref{alg:odd}. Note that the ODD score is an image-level measurement. Therefore, the algorithm will return the detection difficulty for one image. The algorithm's input is the same as calculating mAP, which includes prediction results (PredBbox, PredLabel with its confidence) and ground truth (GtBbox and GtLabel). In addition, the algorithm has two hyperparameters: near-positive threshold ($t_1$) and positive threshold ($t_2$). These two parameters are used to differentiate positive, near-positive, and multi-positive samples, which have been mentioned in Section 3.1 of our paper. After running the algorithm, we can obtain the corresponding ODD score.

\section{Effect of ODD Strategy for Global-frame-based VOD}

Table \ref{tab:odd_strategy} shows three VOD models with different ODD strategies. Recall that ODD-A uses ODD Scheduler with ODD Threshold for faster detection speed and ODD-B/C uses ODD-based Global Reference Frame Selector (OGRFS) for better detection accuracy in both training and inference stages. Note that using ODD-B/C will not reduce the detection speed. Therefore, the ODD-full version (ODD-A+B+C) ODD-VOD can get the best speed-accuracy trade-off but need to retrain the VOD model. ODD-A+B strategy can also get a good trade-off without retraining VOD models, which provides an additional option for how ODD-VOD can be used.

\begin{table}[htbp]
\centering
\caption{Results for Global-frame-based VOD with different ODD strategies. We use ResNet-50 as the backbone for all models.}
\label{tab:odd_strategy}
\resizebox{0.5\textwidth}{!}
{
\begin{tabular}{c|l|ccccccc|c|c}
\toprule[1pt]
\multirow{2}{*}{\textbf{Model}} & \multicolumn{1}{c|}{\multirow{2}{*}{\textbf{\begin{tabular}[c]{@{}c@{}}ODD \\ Strategy\end{tabular}}}} & \multirow{2}{*}{\textbf{}} & \multicolumn{6}{c|}{ODD Theshold} & \multirow{2}{*}{\textbf{\begin{tabular}[c]{@{}c@{}}Original\\ Results\end{tabular}}} & \multirow{2}{*}{\textbf{\begin{tabular}[c]{@{}c@{}}Lossless\\ Acc. Rate\end{tabular}}} \\
 & \multicolumn{1}{c|}{} &  & \textbf{0.8} & \textbf{0.4} & \multicolumn{1}{c}{\textbf{0.3}} & \textbf{0.2} & \textbf{0.15} & \textbf{0.1}  &  &  \\ \hline
\multirow{4}{*}{\begin{tabular}[c]{@{}c@{}}SELSA\\ \cite{selsa}\end{tabular}} & A & \multicolumn{1}{c|}{mAP} & 73.5 & 76.3 & 77.2 & 78.1 & 78.2 & \textbf{78.4} & \multirow{3}{*}{78.4} & 32.8\% \\
 & A+B & \multicolumn{1}{c|}{mAP} & 74.2 & 77.4 & \textbf{78.5} & \textbf{79.2} & \textbf{79.4} & \textbf{79.6}  &  & 72.4\% \\
 & A+B+C & \multicolumn{1}{c|}{mAP} & 74.4 & 77.6 & \textbf{78.7} & \textbf{79.5} & \textbf{79.7} & \textbf{79.8} &  & 72.4\% \\ \cline{2-11} 
 &  & \multicolumn{1}{l|}{FPS} & 33.7 & 22.7 & 20.0 & 18.1 & 16.6 & 15.4 & 11.6 & - \\ \hline
\multirow{4}{*}{\begin{tabular}[c]{@{}c@{}}TROIA\\ \cite{temporal_roi_align}\end{tabular}} & A & \multicolumn{1}{c|}{mAP} & 73.8 & 77.5 & 78.2 & 79 & 79.4 & 79.6  & \multirow{3}{*}{79.8} & 15\% \\
 & A+B & \multicolumn{1}{c|}{mAP} & 75.4 & 78.3 & 79.0 & 79.7 & \textbf{79.9} & \textbf{80}  &  & 40\% \\
 & A+B+C & \multicolumn{1}{c|}{mAP} & 75.5 & 78.8 & 79.7 & \textbf{80.1} & \textbf{80.2} & \textbf{80.4}  &  & 51.7\% \\ \cline{2-11} 
 &  & \multicolumn{1}{l|}{FPS} & 24.6 & 13.0 & 10.8 & 9.1 & 8.4 & 7.6 & 6.0 & - \\ \hline
\multirow{4}{*}{\begin{tabular}[c]{@{}c@{}}MEGA\\ \cite{mega}\end{tabular}} & A & \multicolumn{1}{c|}{mAP} & 72.5 & 75.4 & 76.2 & 76.8 & \textbf{77} & \textbf{77} & \multirow{3}{*}{77.0} & 70.8\% \\
 & A+B & \multicolumn{1}{c|}{mAP} & 72.9 & 76 & \textbf{77.0} & \textbf{77.8} & \textbf{78} & \textbf{78.2} &  & 112.5\% \\
 & A+B+C & \multicolumn{1}{c|}{mAP} & 73 & 76.1 & \textbf{77.2} & \textbf{77.9} & \textbf{78.2} & \textbf{78.4} &  & 112.5\% \\ \cline{2-11} 
 &  & \multicolumn{1}{l|}{FPS} & 21.9 & 12.4 & 10.2 & 8.7 & 8.2 & 7.7  & 4.8 & - \\ 
 \bottomrule[1pt]
\end{tabular}
}
\end{table}

\end{document}